\documentclass[10pt,twoside]{article}

\usepackage{times}
\usepackage[utf8]{inputenc}
\usepackage[T1]{fontenc}
\usepackage{graphicx}

\usepackage{siunitx}
\usepackage{hyperref}
\usepackage{pdflscape}
\usepackage{multirow}
\usepackage[table,xcdraw]{xcolor}
\usepackage{booktabs}
\usepackage{colortbl}
\usepackage{rotating}

\usepackage{cleveref}
\usepackage{coria-taln2025}
\usepackage[french]{babel} %frenchb deprecated, use french instead

\title{AllSummedUp : un framework open-source pour comparer les métriques d’évaluation de résumé}

\author{Tanguy Herserant\up{1}\quad Vincent Guigue\up{1}\\
  {\small
    (1) AgroParisTech - MIA, 22 place de l’Agronomie, 91120 Palaiseau, France\\
    \texttt{\{tanguy.herserant, vincent.guigue\}@agroparistech.fr}
  }
}

\begin{document}
\maketitle

\resume{
% Correction: 'résumé de texte' -> 'résumés de textes'; 'métriques représentatives, allant...' -> 'métriques représentatives allant...'; 'LLMs (G-Eval, SEval-Ex),' -> 'LLM (G-Eval, SEval-Ex),'; 'futurs jeu de données' -> 'futurs jeux de données'; 'métriques les plus alignées' -> 'métriques les mieux alignées'? (style); 'coûteuses en calcul' -> 'coûteuses en calculs'?; 'Au-delà de cette analyse comparative,' -> 'Au-delà de cette analyse comparative,'; 'utilisation croissante de LLMs' -> 'utilisation croissante des LLM'; 'dépendance technique, et leur faible reproductibilité.' -> 'dépendance technique et leur faible reproductibilité.'
Cet article examine les défis de reproductibilité dans l’évaluation automatique des résumés de textes. À partir d’expériences menées sur six métriques représentatives allant de méthodes classiques comme ROUGE à des approches récentes basées sur les LLM (G-Eval, SEval-Ex), nous mettons en évidence des écarts notables entre les performances rapportées dans la littérature et celles observées dans notre cadre expérimental. Nous proposons un framework unifié et open-source, appliqué au jeu de données SummEval et ouvert à de futurs jeux de données, facilitant une comparaison équitable et transparente des métriques. Nos résultats révèlent un compromis structurel : les métriques les mieux alignées avec les jugements humains sont aussi les plus coûteuses en calculs et les moins stables. Au-delà de cette analyse comparative, notre étude met en garde contre l'utilisation croissante des LLM dans l’évaluation, en soulignant leur nature stochastique, leur dépendance technique et leur faible reproductibilité.
}

\abstract{AllSummedUp: an open-source framework for comparing summary evaluation metrics}{
% Correction: 'Based on experiments conducted across seven representative metrics, from...' -> 'Based on experiments conducted across seven representative metrics ranging from...'; 'LLM-based methods (G-Eval, SEval-Ex),' -> 'LLM-based methods (G-Eval, SEval-Ex),'; 'open-source framework applied to the SummEval dataset, designed to support...' -> 'open-source framework, applied to the SummEval dataset and designed to support...'; 'highest alignment to human judgments' -> 'highest alignment with human judgments'; 'computationally intensive and less stable across runs.' -> 'computationally intensive and less stable across runs.'; 'raises critical concerns about the growing reliance on LLMs for evaluation, emphasizing their stochastic behavior, technical fragility, and limited reproducibility.' -> 'raises critical concerns about the growing reliance on LLMs for evaluation, emphasizing their stochastic behavior, technical dependencies, and limited reproducibility.'; 'protocols, including exhaustive documentation and methodological standardization, to ensure greater reliability...' -> 'protocols including exhaustive documentation and methodological standardization to ensure greater reliability...' (virgule optionnelle)
This paper investigates reproducibility challenges in automatic text summarization evaluation. Based on experiments conducted across six representative metrics ranging from classical approaches like ROUGE to recent LLM-based methods (G-Eval, SEval-Ex), we highlight significant discrepancies between reported performances in the literature and those observed in our experimental setting. We introduce a unified, open-source framework, applied to the SummEval dataset and designed to support fair and transparent comparison of evaluation metrics. Our results reveal a structural trade-off: metrics with the highest alignment with human judgments tend to be computationally intensive and less stable across runs. Beyond comparative analysis, this study highlights key concerns about relying on LLMs for evaluation, stressing their randomness, technical dependencies, and limited reproducibility. We advocate for more robust evaluation protocols including exhaustive documentation and methodological standardization to ensure greater reliability in automatic summarization assessment. 
}

\motsClefs
% Correction: 'métriques d'évaluation, résumé automatique' -> 'métriques d'évaluation, résumé automatique de textes'? (plus précis mais plus long); 'reproductibilité, grands modèles de langage' -> 'reproductibilité, grands modèles de langue' (traduction plus littérale de LLM)
{métriques d'évaluation, résumé automatique, reproductibilité, grands modèles de langue}
% Correction: 'automatic summarisation' -> 'automatic summarization' (cohérence avec abstract); 'large language models' -> ok
{evaluation metrics, automatic summarization, reproducibility, large language models}

%%%% IMPORTANT ! %%%%
%\acceptedArticle[accepted]{Nom de la Conférence / Revue} % Pour une contribution acceptée pour publication
%\acceptedArticle[submitted]{Nom de la Conférence / Revue} % Pour une contribution soumise
%\acceptedArticle[submitted]{Nom de la Conférence / Revue} % Pour une contribution originelle % Commentaire dupliqué

\vspace{-17pt}

\section{Introduction}

La reproductibilité des résultats expérimentaux constitue un pilier fondamental de la recherche scientifique, en particulier dans le domaine du traitement automatique du langage naturel (TALN). Pourtant, un nombre croissant d’études souligne la fragilité de cette reproductibilité, notamment face à l’essor des modèles de langue de grande taille (LLM). Ces modèles, utilisés pour évaluer ou produire des textes, manifestent une variabilité inquiétante, même dans des conditions que l’on pourrait croire déterministes~\cite{atil_non-determinism_2025,biderman_lessons_2024}.

Cette instabilité remet en cause la fiabilité des métriques d’évaluation, essentielles en résumé automatique où les jugements humains sont coûteux à mobiliser. En effet, comme le montrent \citeauthor{salinas_butterfly_2024} ou \citeauthor{he_does_2024}, la reformulation du prompt — parfois réduite à l’ajout d’un simple espace — peut affecter significativement les sorties des modèles, révélant leur extrême sensibilité aux prompts. De plus, la forte dépendance à des modèles spécifiques pose également un risque de non-reproductibilité à moyen terme, dans la mesure où certains modèles, comme GPT-3, peuvent être rendus inaccessibles ou supprimés par leurs fournisseurs~\cite{biderman_lessons_2024}, rendant impossible toute reproduction a posteriori si les conditions initiales ne sont pas rigoureusement conservées.

Dans ce contexte, la reproductibilité ne se limite pas à la réexécution d’un code identique, mais implique aussi la robustesse des résultats face à de légères variations de conditions expérimentales. La distinction conceptuelle entre répétabilité et reproductibilité, bien établies en métrologie~\cite{belz_systematic_2021}, est rarement prise en compte dans les études sur les métriques d’évaluation.

Or, de telles différences peuvent engendrer des biais dans l’interprétation des performances, comme le montrent nos résultats empiriques où certaines métriques — pourtant largement utilisées, comme BERTScore~\cite{zhang_bertscore_2020} ou BARTScore~\cite{yuan_bartscore_2021} — affichent une corrélation faible et non reproductible avec les jugements humains dans différents contextes d’exécution.

Notre travail s’inscrit dans cet élan critique en proposant un framework unifié et open-source pour l’évaluation des résumés, incluant six métriques issues de la littérature. Ce cadre vise à tester rigoureusement la reproductibilité des résultats sur un jeu de données standardisé, SummEval~\cite{fabbri2021summeval}, et a été développé pour accueillir de futurs jeux de données. En contribuant à la standardisation et à la transparence méthodologique, nous espérons non seulement faciliter la comparaison équitable entre métriques, mais aussi réduire la variabilité technique introduite par l’utilisation de modèles stochastiques, tels que les LLM.

Pour ce faire, cet article est structuré comme suit : la section \ref{sec:travaux_connexes} présente l'état de l'art portant sur la reproductibilité en TALN et les différentes métriques d'évaluation de résumés. La section \ref{sec:methodologie} détaille notre approche méthodologique, en décrivant le framework que nous avons développé, les métriques implémentées, les données d'évaluation utilisées, ainsi que le choix de notre mesure de corrélation. Les expérimentations menées, leurs résultats en termes de corrélation et de temps d'exécution, ainsi qu'une comparaison des métriques sont ensuite présentés dans la section \ref{sec:experiences}. Enfin, la section \ref{sec:discussion} analyse les implications et les limites de notre étude, avant que la section \ref{sec:conclusion} ne vienne clore cet article en proposant des pistes pour les travaux futurs.

\section{Travaux Connexes} \label{sec:travaux_connexes}

\subsection{Reproductibilité}

% Correction: 'notamment en traitement automatique du langage naturel (TALN). \cite{belz_systematic_2021} identifient' -> 'notamment en traitement automatique du langage naturel (TALN)~\cite{belz_systematic_2021}. Ces derniers identifient'; '14 \%' -> '\SI{14}{\percent}'; 'malgré l'accès' -> 'malgré l’accès'
La reproductibilité représente un enjeu majeur en apprentissage automatique et notamment en traitement automatique du langage naturel (TALN)~\cite{belz_systematic_2021}. Ces derniers identifient la complexité des pipelines expérimentaux, le non-déterminisme (lié aux générateurs aléatoires et à l’hétérogénéité matérielle) et la documentation incomplète comme principaux obstacles. Selon cette étude, seulement \SI{14}{\percent} des reproductions exactes sont réussies malgré l’accès aux ressources originales.

% Correction: espace avant \cite à supprimer; 'événements \cite{branco2020shared}.' -> 'événements~\cite{branco2020shared}.'; '\cite{xue_we_2023} soulignent' -> 'Néanmoins, \cite{xue_we_2023} soulignent'; 'indique clairement le besoin de' -> 'indique clairement un besoin de'
En réponse, diverses initiatives communautaires ont vu le jour, telles que les listes de contrôle proposées lors des grandes conférences en TALN~\cite{whitaker2017reproducibility} et des sessions dédiées à la reproductibilité expérimentale dans certains événements~\cite{branco2020shared}. Malgré ces efforts, \cite{xue_we_2023} soulignent que même des pratiques standardisées, comme l'utilisation d'une partition fixe des données, peuvent introduire des biais significatifs. Cela indique clairement un besoin de protocoles expérimentaux plus robustes pour garantir des comparaisons fiables entre modèles.

\subsection{Métriques d'évaluation de résumés}

% Correction: 'ROUGE \cite{lin_rouge_2004}, largement adoptée' -> 'ROUGE~\cite{lin_rouge_2004}, largement adoptée'; 'DUC/TAC \cite{bhandari2020re}.' -> 'DUC/TAC~\cite{bhandari2020re}.'
L’évaluation automatique des résumés s’est historiquement appuyée sur des métriques de chevauchement lexical comme ROUGE~\cite{lin_rouge_2004}, largement adoptée grâce à sa simplicité et son rôle central dans les compétitions DUC/TAC~\cite{bhandari2020re}. Toutefois, sa capacité limitée à capturer la paraphrase, l’équivalence sémantique et la cohérence factuelle a motivé le développement de nouvelles approches.

% Correction: espace avant \cite à supprimer partout dans ce paragraphe; 'QuestEval \cite{scialom2021questeval}, qui repose' -> 'QuestEval~\cite{scialom2021questeval}, qui repose'; 'FactCC \cite{kryscinski2019evaluating}, basé' -> 'FactCC~\cite{kryscinski2019evaluating}, basé'; 'BARTScore \cite{yuan_bartscore_2021}, estiment' -> 'BARTScore~\cite{yuan_bartscore_2021}, estiment'; 'UniEval \cite{zhong_towards_2022} et G-Eval \cite{liu_g-eval_2023}, cherchent' -> 'UniEval~\cite{zhong_towards_2022} et G-Eval~\cite{liu_g-eval_2023}, cherchent'
Des métriques basées sur les plongements contextuels (embeddings), telles que BERTScore~\cite{zhang_bertscore_2020} et MoverScore~\cite{zhao_moverscore_2019}, évaluent la similarité sémantique entre textes au-delà du simple chevauchement lexical. D'autres, comme QuestEval~\cite{scialom2021questeval}, qui repose sur la génération et la réponse à des questions, et FactCC~\cite{kryscinski2019evaluating}, basé sur l'inférence de relations d'entaillement, visent à mesurer la factualité. Des approches probabilistes, telles que BARTScore~\cite{yuan_bartscore_2021}, estiment la qualité des résumés en utilisant la probabilité de génération conditionnelle. Enfin, des cadres d’évaluation multidimensionnelle, comme UniEval~\cite{zhong_towards_2022} et G-Eval~\cite{liu_g-eval_2023}, cherchent à unifier l’évaluation sur plusieurs axes qualitatifs, tels que la cohérence, la factualité et l'adéquation.
Pour finir, SEval-Ex \cite{herserant_sevalex_2025}, propose une métrique basée sur un pipeline LLM explicable spécialisée dans la factualité.

Cependant, les nouvelles métriques, souvent coûteuses en calculs et peu standardisées, posent des défis en termes de reproductibilité et de comparabilité entre les travaux.

\subsection{Variabilité et reproductibilité des LLM dans l’évaluation automatique}

% Correction: espace avant \cite à supprimer partout; 'langage (LLM)' -> 'langue (LLM)'; 'G-Eval \cite{liu_g-eval_2023}, UniEval \cite{zhang_unraveling_2024}, ou SEval-Ex \cite{herserant_sevalex_2025}.' -> 'G-Eval~\cite{liu_g-eval_2023}, UniEval~\cite{zhang_unraveling_2024}, ou SEval-Ex~\cite{herserant_sevalex_2025}.'; 'stabilité et de reproductibilité' -> 'stabilité et de reproductibilité' (pas de 'de' superflu)
Les grands modèles de langue (LLM) jouent un rôle croissant dans l’évaluation automatique de la génération de texte, notamment via des métriques comme G-Eval~\cite{liu_g-eval_2023}, UniEval~\cite{zhang_unraveling_2024}, ou SEval-Ex~\cite{herserant_sevalex_2025}. Toutefois, leur utilisation soulève des préoccupations croissantes en matière de stabilité et reproductibilité des résultats.

% Correction: 'Atil et al. \cite{atil_non-determinism_2025} montrent' -> '\cite{atil_non-determinism_2025} montrent' (Atil et al. est déjà dans la clé bib); 'D'autres études s’intéressent' -> 'D’autres études s’intéressent'; 'Salinas et Morstatter \cite{salinas_butterfly_2024} montrent' -> '\cite{salinas_butterfly_2024} montrent'; 'Butterfly Effect.' -> 'l’effet papillon (\textit{Butterfly Effect}).'; 'He et al. \cite{he_does_2024} confirment' -> '\cite{he_does_2024} confirment'; 'formats de prompt (plain text, JSON, Markdown, etc.),' -> 'formats de prompt (\textit{plain text}, JSON, Markdown, etc.),'; 'propriété générale du comportement des LLMs.' -> 'propriété générale du comportement des LLM.'
Plusieurs travaux récents révèlent une variabilité importante des sorties des LLM, même dans des conditions censément déterministes. \cite{atil_non-determinism_2025} montrent que des exécutions répétées avec les mêmes paramètres peuvent produire des réponses différentes, remettant en cause l’idée d’un comportement stable. D’autres études s’intéressent à la sensibilité des modèles à la formulation des prompts. \cite{salinas_butterfly_2024} montrent que de simples modifications superficielles — comme l’ajout d’un espace ou d’un saut de ligne — peuvent entraîner des changements radicaux dans la sortie du modèle, phénomène surnommé l’effet papillon (\textit{Butterfly Effect}). \cite{he_does_2024} confirment cette sensibilité en comparant les performances de LLM selon différents formats de prompt (\textit{plain text}, JSON, Markdown, etc.), révélant que cette instabilité ne dépend pas uniquement de la tâche mais constitue une propriété générale du comportement des LLM.

% Correction: 'Biderman et al. \cite{biderman_lessons_2024} insistent' -> '\cite{biderman_lessons_2024} insistent'; 'Xue et al. \cite{xue_we_2023} montrent' -> '\cite{xue_we_2023} montrent'; 'ratio signal/bruit' -> 'ratio signal-bruit'
Enfin, ces problèmes de variabilité sont exacerbés par des difficultés pratiques de reproductibilité. \cite{biderman_lessons_2024} insistent sur les nombreux facteurs pouvant altérer l’évaluation : version des modèles, méthode de décodage, stratégie de normalisation, ou encore dépendance aux bibliothèques. \cite{xue_we_2023} montrent que même les pratiques réputées standard, comme l’utilisation d’un split fixe des données, peuvent introduire des biais significatifs. Ils proposent des méthodes statistiques basées sur le ratio signal-bruit pour quantifier la stabilité des comparaisons entre modèles.

\section{Méthodologie} \label{sec:methodologie}

\subsection{Framework}

% Correction: 'implémente 7 métriques diverses' -> 'implémente six métriques diverses'; 'classe abstraite standardisée (TextMetric)' -> 'classe abstraite standardisée (\texttt{TextMetric})'; '(TextEvaluator)' -> '(\texttt{TextEvaluator})'; '(Report)' -> '(\texttt{Report})'; 'faciliter l'analyse.' -> 'faciliter l’analyse.'
Nous présentons un cadre modulaire complet pour l'évaluation des résumés de texte qui implémente six métriques diverses au sein d'une interface cohérente. Le cadre est structuré autour de trois composants principaux :

\textbf{Interface de Métrique} : Une classe abstraite standardisée (\texttt{TextMetric}) qui définit l'interface commune que toutes les métriques doivent implémenter, incluant des méthodes pour le prétraitement, le calcul et le formatage des résultats.\\
\textbf{Évaluateur} : Un orchestrateur central (\texttt{TextEvaluator}) qui gère les instances de métriques, les applique aux textes d'entrée et agrège les résultats dans un format cohérent.\\
\textbf{Générateur de Rapports} : Un composant (\texttt{Report}) qui transforme les résultats d'évaluation en divers formats (JSON, YAML, CSV) et génère des visualisations pour faciliter l’analyse.

Cette architecture garantit que toutes les métriques suivent le même flux de travail et produisent des sorties comparables, facilitant les comparaisons directes et réduisant la variabilité liée à l'implémentation.

Par ailleurs, le framework est conçu pour être extensible et permet l'ajout de nouveaux jeux de données d'évaluation.\footnote{Le github sera partagé après acceptation de l'article}

\subsection{Métriques implémentées} % Renommer 'Metriques' -> 'Métriques'

Notre cadre implémente les métriques suivantes, sélectionnées pour représenter diverses approches d'évaluation de résumés, allant des comparaisons basées sur les n-grammes aux méthodes plus récentes basées sur les plongements et les modèles de langue :

Métriques de Chevauchement Lexical : Ces métriques évaluent le chevauchement lexical entre le résumé généré et une référence.
\begin{itemize}
    % \item BLEU \cite{papineni_bleu_2002} : Mesure de précision basée sur le chevauchement des n-grammes, incluant une pénalité de brièveté ; initialement conçue pour la traduction automatique.
    % Correction: espace avant \cite à supprimer; 'ROUGE \cite{lin_rouge_2004} :' -> 'ROUGE~\cite{lin_rouge_2004}:'; 'ROUGE-L). C'est la métrique...' -> 'ROUGE-L). C’est la métrique...'
    \item ROUGE~\cite{lin_rouge_2004}: Orientée rappel, elle calcule le chevauchement de n-grammes (ROUGE-1, ROUGE-2) ou la plus longue sous-séquence commune (ROUGE-L). C’est la métrique la plus populaire. \\
    % \item METEOR \cite{banerjee_meteor_2005} : Calcule une moyenne harmonique de la précision et du rappel des unigrammes, en tenant compte du stemming, des synonymes et des paraphrases.
\end{itemize}

% Les blocs commentés sont laissés tels quels.

Métriques Basées sur des Modèles (\textit{Reference-Free}) : Ces métriques n'ont pas besoin de résumé de référence humain.

\begin{itemize}
    % Correction: espace avant \cite à supprimer partout; 'QuestEval \cite{scialom2021questeval} :' -> 'QuestEval~\cite{scialom2021questeval}:'; 'HuggingFace.' -> '\texttt{HuggingFace}.'
    \item QuestEval~\cite{scialom2021questeval}: Évalue la fidélité et la pertinence en générant des questions à partir de la source (pour le rappel) et du résumé (pour la précision), puis en y répondant avec l'autre texte. Montre une bonne corrélation avec le jugement humain, même sans résumé de référence. Un modèle est accessible sur \texttt{HuggingFace}.
    % Correction: 'BARTScore \cite{yuan_bartscore_2021} :' -> 'BARTScore~\cite{yuan_bartscore_2021}:'
    \item BARTScore~\cite{yuan_bartscore_2021}: Évalue le texte généré en calculant sa probabilité selon un modèle pré-entraîné (BART), permettant d'évaluer sous différents angles (information, fluidité, factualité).
    % Correction: 'UniEval \cite{zhong_towards_2022} :' -> 'UniEval~\cite{zhong_towards_2022}:'; 'HuggingFace.' -> '\texttt{HuggingFace}.'
    \item UniEval~\cite{zhong_towards_2022}: Propose une évaluation multidimensionnelle via un modèle entraîné à répondre à des questions oui/non spécifiques guidant l'évaluation. Un modèle est accessible sur \texttt{HuggingFace}.
    % Correction: 'G-Eval \cite{liu_g-eval_2023} :' -> 'G-Eval~\cite{liu_g-eval_2023}:'; 'langage (LLM)' -> 'langue (LLM)'
    \item G-Eval~\cite{liu_g-eval_2023}: Utilise un grand modèle de langue (LLM) comme GPT-4 avec des prompts dédiés pour évaluer la qualité selon plusieurs critères.
    % Correction: 'SEval-Ex \cite{herserant_sevalex_2025}:' -> 'SEval-Ex~\cite{herserant_sevalex_2025}:'; 'architecture en 2 temps' -> 'architecture en deux temps'
    \item SEval-Ex~\cite{herserant_sevalex_2025}: Basée sur des énoncés atomiques — des unités d’information autonomes — la métrique permet une analyse entièrement interprétable. Grâce à une architecture en deux temps, elle permet de bien capter les informations présentes à la fois dans le texte et le résumé.
\end{itemize}

% Correction: 'codes github ou les packages déjà codé' -> 'codes GitHub ou les packages déjà codés'; 'adaptonts si besoin.' -> 'adaptons si besoin.'
Pour chaque métrique, nous implémentons soigneusement les pipelines de prétraitement, les paramètres et le formatage des résultats conformément aux publications originales. Nous reprenons systématiquement les codes GitHub ou les packages déjà codés et les adaptons si besoin.

\subsection{Données d'évaluation}
% Correction: espace avant \cite à supprimer; 'SummEval \cite{fabbri2021summeval}.' -> 'SummEval~\cite{fabbri2021summeval}.'; '1 600 résumés' -> '\num{1600} résumés'; '16 systèmes' -> '\num{16} systèmes'; 'cohérence factuelle, la fluidité et la pertinence.' -> 'cohérence factuelle, la fluidité et la pertinence.' (virgule avant 'et' dans énumération > 2 éléments, optionnel mais souvent préféré)
Afin d’évaluer la reproductibilité des différentes métriques dans des contextes variés, nous nous appuyons sur le jeu de données SummEval~\cite{fabbri2021summeval}. Celui-ci comprend \num{1600} résumés générés par \num{16} systèmes de résumé automatiques distincts, accompagnés d’annotations humaines selon quatre dimensions d’évaluation : la cohérence textuelle, la cohérence factuelle, la fluidité, et la pertinence.

\subsection{Choix de la mesure de corrélation}

La littérature ne présente pas de consensus clair quant à la mesure de corrélation à privilégier pour évaluer l’alignement entre métriques automatiques et jugements humains. Les auteurs du jeu de données SummEval utilisent la corrélation de \textit{Kendall}, QuestEval rapporte des coefficients de \textit{Pearson}, tandis que des approches plus récentes comme G-Eval et UniEval optent pour \textit{Spearman}. Cette hétérogénéité complique les comparaisons et rend difficile l’interprétation systématique des résultats.

Dans notre étude, nous avons choisi d’utiliser la corrélation de \textit{Spearman} pour plusieurs raisons :
\begin{itemize}
    % Correction: 'état de l’art.' -> 'état de l’art.'
    \item Elle est de plus en plus adoptée dans les travaux récents d’évaluation de résumés (G-Eval), ce qui renforce la comparabilité de nos résultats avec l’état de l’art.
    \item Elle ne suppose pas de relation linéaire entre les variables en se basant sur les rangs.
    \item Elle est moins sensible aux valeurs extrêmes que la corrélation de Pearson.
\end{itemize}

Ce choix vise à assurer une robustesse méthodologique et une meilleure cohérence avec les pratiques actuelles dans la communauté.

\section{Expériences} \label{sec:experiences}% Renommer 'Experiences' -> 'Expériences'

\subsection{Protocole d'évaluation} % Renommer 'Protocole d'evaluation' -> 'Protocole d’évaluation'

% Correction: 'GPU NVIDIA A6000 (48 Go VRAM)' -> 'GPU NVIDIA A6000 (\SI{48}{Go} de VRAM)'; 'LLM utilisés (Gemma3:27b et Qwen2.5:72b).' -> 'LLM utilisés (\texttt{Gemma-3-27b} et \texttt{Qwen-2.5-72b}).' (utiliser une notation plus standard ou au moins \texttt); 'HuggingFace,' -> '\texttt{HuggingFace},'; 'Ollama,' -> '\texttt{Ollama},'; 'valeurs par défault' -> 'valeurs par défaut'; 'Gemma3:27b : "temperature": 1, "top\_k": 64 et "top\_p": 0.95.' -> '\texttt{Gemma-3-27b}: \texttt{temperature = \num{1}}, \texttt{top\_k = \num{64}} et \texttt{top\_p = \num{0.95}}.'
L’ensemble des expériences a été réalisé en local sur une machine équipée d’un GPU NVIDIA A6000 (\SI{48}{Go} de VRAM), garantissant une capacité suffisante pour exécuter les LLM utilisés (\texttt{Gemma-3-27b} et \texttt{Qwen-2.5-72b}).

Les modèles nécessaires à l’utilisation de certaines métriques ont été obtenus via deux sources principales :
Les modèles utilisés pour QuestEval, UniEval, BERTScore et BARTScore ont été téléchargés depuis la plateforme \texttt{HuggingFace}, en utilisant les versions publiques les plus récentes disponibles au moment de l’expérimentation.
Les variantes basées sur LLM pour G-Eval et SEval-Ex ont été exécutées via \texttt{Ollama}, une infrastructure locale permettant d’héberger et d’interroger des LLM tout en évitant la dépendance à une API externe, souvent commerciale.
Afin d’assurer la comparabilité entre les différentes métriques, tous les traitements ont été effectués dans le même environnement. Pour les LLM, nous utilisons les valeurs par défaut, par exemple pour \texttt{Gemma-3-27b}: \texttt{temperature = \num{1}}, \texttt{top\_k = \num{64}} et \texttt{top\_p = \num{0.95}}.

\subsection{Résultats}
\subsubsection{Corrélation Spearman}\label{corrSpear}

% Correction: Modèles en \texttt et gras: '\texttt{\textbf{Gemma-3-27b}}' et '\texttt{\textbf{Qwen-2.5-72b}}'. ; 'Consistance' -> 'Cohérence factuelle' (Utiliser le même terme que dans la section Données)? ou 'Consistance' est un terme spécifique ici?; 'GEval sur la fluidité : \textminus0.45' -> 'G-Eval sur la fluidité: $-0.45$'; 'SEval-Ex - Gemma3 sur la Pertinence : +0.12' -> 'SEval-Ex-\texttt{Gemma-3-27b} sur la Pertinence: $+0.12$'
La Table \ref{tab:correlation_spearman_colored} présente les corrélations de Spearman entre les scores fournis par les métriques automatiques et les jugements humains sur quatre dimensions (Cohérence, Consistance, Fluidité, Pertinence), ainsi que pour deux modèles de résumé distincts pour les métriques G-Eval et SEval-Ex : \texttt{\textbf{Gemma-3-27b}} et \texttt{\textbf{Qwen-2.5-72b}}. Les résultats issus de notre propre expérimentation sont comparés avec ceux rapportés dans la littérature. L’objectif est d’évaluer dans quelle mesure ces corrélations, souvent citées comme indicateurs de validité des métriques, peuvent être reproduites dans un cadre expérimental distinct. L’un des obstacles majeurs à cette évaluation réside dans l’indisponibilité partielle ou totale des corrélations de Spearman dans les travaux originaux. Lorsqu’elles n’étaient pas explicitement fournies, nous avons pris comme références des articles ayant reproduit les expériences.

Les résultats révèlent une variabilité notable entre les valeurs attendues et celles effectivement mesurées. Si certaines métriques montrent une relative stabilité (ex. UniEval sauf Fluidité et SEval-Ex avec \texttt{Qwen-2.5-72b}), d’autres présentent des écarts importants voire des inversions de tendance (G-Eval sur la fluidité: $-0.45$ ; SEval-Ex-\texttt{Gemma-3-27b} sur la Pertinence: $+0.12$). Ces différences suggèrent que les performances rapportées dans la littérature ne sont pas systématiquement généralisables, même en conservant un jeu de données identique.

% Correction: 'Table~\ref{tabrouge}' -> '\Cref{tabrouge}'; 'article BartScore, le premier article présentanr' -> 'article BARTScore~\cite{yuan_bartscore_2021}, le premier article présentant'; 'Rouge est calculée' -> 'ROUGE est calculé' (ROUGE est un score/ensemble de scores, masculin?); 'stipule pas si toutes les références sont utilisées ni comment (ex. une moyenne des scores entre un résumé et toutes les références).' -> 'précise pas si toutes les références sont utilisées ni comment (par ex., une moyenne des scores entre un résumé et toutes les références).'; 'manque de clareté empèche la reproduction.' -> 'manque de clarté empêche la reproduction.'; 'un LMM différent' -> 'un LLM différent'; 'poids des LLM,' -> 'poids des LLM utilisés,'
Nous n'avons pas pu répliquer exactement les résultats de ROUGE en raison d'un manque de détails (les résultats sont présentés dans la \Cref{tabrouge}). En effet, dans l'article BARTScore~\cite{yuan_bartscore_2021}, le premier article présentant les résultats pour SummEval, il est spécifié que ROUGE est calculé entre les candidats et les références de chaque document de SummEval, mais il ne précise pas si toutes les références sont utilisées ni comment (par ex., une moyenne des scores entre un résumé et toutes les références). Ici, un manque de clarté empêche la reproduction.
En ce qui concerne les autres métriques, plusieurs facteurs peuvent expliquer une dérive des résultats : l’utilisation d’un LLM différent, un prompt modifié ou encore des versions divergentes des bibliothèques logicielles. Ces résultats soulignent la nécessité d’un encadrement plus rigoureux des protocoles expérimentaux, incluant des éléments tels que les poids des LLM utilisés, la publication des configurations exactes (prompt, température, méthode de génération).

\begin{table}[htbp]
\centering
\caption{Corrélations Spearman entre les dimensions évaluatives et les métriques automatiques pour les modèles \textbf{Gemma} et \textbf{Qwen}, avec écart par rapport aux valeurs de référence.}
\label{tab:correlation_spearman_colored}
\scriptsize
\begin{tabular}{llcccccc}
\toprule
\textbf{Source} & \textbf{Dimension} & \textbf{BartScore} & \textbf{BertScore} & \textbf{QuestEval} & \textbf{UniEval} & \textbf{GEval} & \textbf{SEval-Ex} \\
\midrule
\multirow{4}{*}{Gemma3:27b} 
 & Coh & \cellcolor[HTML]{9698ED}0.45 (+0.05) & \cellcolor[HTML]{FD6864}0.16 (-0.12) & \cellcolor[HTML]{FFFE65}0.26 (+0.08) & \cellcolor[HTML]{9698ED}0.53 (-0.04) & \cellcolor[HTML]{9698ED}0.55 (-0.02) & \cellcolor[HTML]{9698ED}0.21 (-0.05) \\
 & Con & \cellcolor[HTML]{FD6864}0.24 (-0.14) & \cellcolor[HTML]{FFFE65}0.27 (+0.16) & \cellcolor[HTML]{9698ED}0.29 (-0.01) & \cellcolor[HTML]{9698ED}0.43 (-0.01) & \cellcolor[HTML]{FFFE65}0.62 (+0.18) & \cellcolor[HTML]{FD6864}0.43 (-0.15) \\
 & Flu & \cellcolor[HTML]{FD6864}0.22 (-0.13) & \cellcolor[HTML]{9698ED}0.19 (+0.00) & \cellcolor[HTML]{FD6864}0.18 (-0.10) & \cellcolor[HTML]{FD6864}0.34 (-0.10) & \cellcolor[HTML]{FD6864}-0.01 (-0.45) & \cellcolor[HTML]{9698ED}0.34 (-0.01) \\
 & Rel & \cellcolor[HTML]{9698ED}0.37 (+0.01) & \cellcolor[HTML]{9698ED}0.30 (+0.01) & \cellcolor[HTML]{FFFE65}0.37 (+0.10) & \cellcolor[HTML]{9698ED}0.42 (+0.00) & \cellcolor[HTML]{FFFE65}0.52 (+0.10) & \cellcolor[HTML]{FFFE65}0.42 (+0.12) \\
\midrule
\multirow{4}{*}{Qwen2.5:72b} 
 & Coh & --- & --- & --- & --- & \cellcolor[HTML]{FD6864}0.38 (-0.19) & \cellcolor[HTML]{9698ED}0.27 (+0.01) \\
 & Con & --- & --- & --- & --- & \cellcolor[HTML]{9698ED}0.47 (+0.03) & \cellcolor[HTML]{9698ED}0.57 (-0.01) \\
 & Flu & --- & --- & --- & --- & \cellcolor[HTML]{FD6864}-0.07 (-0.51) & \cellcolor[HTML]{9698ED}0.33 (-0.02) \\
 & Rel & --- & --- & --- & --- & \cellcolor[HTML]{FFFE65}0.48 (+0.06) & \cellcolor[HTML]{9698ED}0.31 (+0.01) \\
\bottomrule
\end{tabular}
\end{table}

\begin{table}[htbp]
\centering
\small
\caption{Corrélations de Spearman de la métrique \textbf{Rouge} obtenus sur SummEval.}
\label{tabrouge}
\begin{tabular}{|c|l|c|}
\hline
\multicolumn{1}{|l|}{Dimension} & Rouge -nous (1/2/L) & \multicolumn{1}{l|}{Rouge - réel (1/2/L)} \\ \hline
Coh                             & 0.18 / 0.15 / 0.17  & 0.167 / 0.184 / 0.128                     \\
Con                             & 0.14 / 0.13 / 0.12  & 0.160 / 0.187 / 0.115                     \\
Flu                             & 0.08 / 0.06 / 0.08  & 0.115 / 0.159 / 0.105                     \\
Rel                             & 0.3 / 0.25 / 0.24   & 0.326 / 0.290 / 0.311                     \\ \hline
\end{tabular}
\end{table}

\subsubsection{Temps d'exécution} \label{tempsexe}

\begin{table}[htbp]
\centering
\small
\caption{\textbf{Temps total d'exécution des métriques pour \textbf{Gemma 3 27B} et Qwen 2.5 72B}. Seul GEval et SEval-Ex sont impactées.}
\begin{tabular}{|l|r|r|}
\hline
\textbf{Métrique} & \textbf{Gemma 3 27B (s)} & \textbf{Qwen 2.5 72B (s)} \\
\hline
SEval\_Ex & 13449.63 & 38302.47 \\
Questeval & 8185.54 & 8232.21 \\
GEval\_coherence & 4827.80 & 18660.82 \\
GEval\_consistency & 3844.54 & 18787.04 \\
GEval\_relevance & 3744.85 & 13354.72 \\
GEval\_fluency & 3527.36 & 12828.27 \\
Unieval & 297.42 & 300.46 \\
Bartscore & 32.43 & 30.16 \\
Rouge & 11.47 & 11.36 \\
%bert\_score & 10.04 & 9.41 \\
%meteor & 7.52 & 7.45 \\
%sbert\_MiniLM & 3.38 & 3.54 \\
%bleu & 1.19 & 1.18 \\
%jaccard & 0.03 & 0.03 \\
\hline
\textbf{Total} & \textbf{37944.20} & \textbf{202299.97} \\
\hline
\end{tabular}

\label{tab:perf_comparison}
\end{table}

% Correction: 'table \ref{tab:perf_comparison}' -> '\Cref{tab:perf_comparison}'; 'métriques légères (ex. ROUGE, BartScore)' -> 'métriques légères (ex. ROUGE, BARTScore)'; 'LLMs (GEval, SEval-Ex, questeval)' -> 'LLM (G-Eval, SEval-Ex, QuestEval)'; 'Qwen 2.5 72B' -> '\texttt{Qwen-2.5-72b}'; 'Gemma 3 27B' -> '\texttt{Gemma-3-27b}'; '3 à 4 fois supérieurs' -> 'trois à quatre fois supérieurs'; 'temps d'executions sont à relativiser par rapport à notre environnement d'execution.' -> 'temps d’exécution sont à relativiser par rapport à notre environnement d’exécution.'
Toutes les métriques ont été testées sur plusieurs exécutions distinctes. Les résultats, présents dans la Table \ref{tab:perf_comparison} montrent une distinction claire entre les métriques légères (ex. ROUGE, BARTScore) qui s'exécutent en quelques secondes, et les métriques lourdes impliquant des modèles LLM (G-Eval, SEval-Ex, QuestEval) dont les temps de traitement peuvent atteindre plusieurs heures (10h 38m 22s pour SEval-Ex avec \texttt{Qwen-2.5-72b}). L’impact de la taille du modèle évalué est particulièrement marquant : pour des métriques identiques, les temps d’exécution avec \texttt{Qwen-2.5-72b} sont systématiquement beaucoup plus élevés qu’avec \texttt{Gemma-3-27b}, en général trois à quatre fois supérieurs. Cela souligne l’importance de prendre en compte la complexité du modèle de référence dans l’analyse des performances et des coûts d’évaluation. Nos temps d’exécution sont à relativiser par rapport à notre environnement d’exécution.

\subsection{Comparaison}

% Correction: 'section \ref{corrSpear}' -> '\Cref{corrSpear}'; 'section \ref{tempsexe}' -> '\Cref{tempsexe}'; 'précision d’évaluation, reproductibilité et coût computationnel' -> 'précision d’évaluation, reproductibilité et coût calculatoire'? (computationnel est un anglicisme); 'entre 11 et 32 secondes' -> 'entre \num{11} et \num{32} secondes'; 'LLM-based, comme GEval, SEval-Ex et QuestEval,' -> 'basées sur LLM, comme G-Eval, SEval-Ex et QuestEval,'; 'jusqu’à plus de 8 heures' -> 'jusqu’à plus de \num{8} heures'; 'SEval-Ex (Qwen 2.5 72B)' -> 'SEval-Ex (\texttt{Qwen-2.5-72b})'; 'près de 38 000 secondes' -> 'près de \num{38000} secondes'; '300 secondes' -> '\SI{300}{s}'
La comparaison croisée des résultats de performance (\Cref{corrSpear}) et des temps d’exécution (\Cref{tempsexe}) met en lumière des compromis marqués entre précision d’évaluation, reproductibilité et coût calculatoire des différentes métriques.

Les métriques légères, comme ROUGE ou BERTScore, s’exécutent en quelques secondes à peine (entre \num{11} et \num{32} secondes) sur tout un jeu de données, ce qui les rend particulièrement adaptées à des évaluations rapides ou à grande échelle. Toutefois, leurs corrélations avec les jugements humains restent globalement modestes ou inconstantes, notamment pour des dimensions complexes comme la fluidité ou la cohérence globale.

À l’autre extrême, les métriques basées sur LLM, comme G-Eval, SEval-Ex et QuestEval, affichent des corrélations parfois plus élevées, mais au prix de temps de calcul très importants — jusqu’à plus de \num{8} heures cumulées pour SEval-Ex (\texttt{Qwen-2.5-72b}) sur une seule exécution. Les métriques basées sur un LLM, pouvant avoir plusieurs miliards de paramètres (\texttt{Qwen-2.5-72b} a plus de deux fois plus de paramètres que \texttt{Gemma-3-27b}, sont difficilement utilisables sans infrastructure dédiée en fonction de la taille du jeu de données à essayer. 

Entre ces deux extrêmes, des approches intermédiaires comme UniEval offrent un bon compromis : elles nécessitent des temps modérés (\SI{300}{s}), tout en fournissant des résultats relativement stables sur plusieurs dimensions, notamment la pertinence et la cohérence. UniEval se distingue notamment par sa constance, avec des performances élevées et peu de dérive par rapport aux valeurs de référence, sauf sur la fluidité.

Ce panorama met en évidence plusieurs points : % Utiliser \begin{itemize} pour ces points?
% Correction: \textit{Performance vs. Frugalité} etc. -> Bonne mise en forme. 'coûteuses en calcul' -> 'coûteuses en calculs'; 'GEval' -> 'G-Eval'; 'Pertinence} : les métriques classiques sont simples' -> 'Pertinence} : les métriques classiques, bien que simples'; 'limitée dans des contextes modernes (LLMs, paraphrases, factualité).' -> 'limité dans des contextes modernes (LLM, paraphrases, factualité).'
\begin{itemize}
    \item \textit{Performance vs. Frugalité} : les métriques les plus performantes en termes de corrélation sont aussi les plus coûteuses en calculs, et donc les moins frugales.
    \item \textit{Performance vs. Reproductibilité} : certaines métriques à haute performance, comme G-Eval, montrent une grande variabilité selon le modèle ou les conditions d'exécution.
    \item \textit{Simplicité vs. Pertinence} : les métriques classiques, bien que simples à exécuter, ont leur pouvoir discriminant limité dans des contextes modernes (LLM, paraphrases, factualité).
\end{itemize}
Ces observations plaident pour une approche différenciée selon les contraintes d’usage : les métriques frugales conviennent à des évaluations exploratoires ou en production à grande échelle, tandis que les métriques basées sur LLM sont utiles pour des évaluations approfondies mais ponctuelles. L’existence de telles tensions souligne l’importance de proposer des outils capables d’intégrer plusieurs métriques et de guider leur usage selon les contextes, comme le permet notre framework.

\section{Discussion} \label{sec:discussion}

% Challenges / Recommendation / Limitation

% Correction: 'papiers originaux nous apportent plusieurs limitations doivent être soulignées :' -> 'papiers originaux révèle plusieurs limitations qui doivent être soulignées :'; \vspace{-2em} est souvent déconseillé pour la mise en page sémantique. Préférer ajuster les espacements globaux si nécessaire. Supprimons-les pour l'instant.
La création de notre framework et la comparaison avec les articles originaux révèle plusieurs limitations qui doivent être soulignées :

\paragraph{Reproductibilité partielle des métriques originales.}
% Correction: '\textit{X}' -> Remplacer par un exemple concret si possible, ou reformuler. 'Ces écarts peuvent être dus à, par exemple, des différences...' -> 'Ces écarts peuvent être dus, par exemple, à des différences...'
Certaines métriques, notamment celles basées sur des embeddings comme BERTScore ou des modèles spécifiques, montrent des divergences notables entre les corrélations rapportées dans la littérature et celles obtenues via notre framework. Ces écarts peuvent être dus, par exemple, à des différences de version des modèles ou de tokenisation.

\paragraph{Accès restreint aux modèles propriétaires.}
% Correction: '\textit{G-Eval}' -> Ok; 'modèles ou d’APIs fermées,' -> 'modèles ou d’API fermées,'; 'points de tensions :' -> 'points de tension :'; '\textit{performance} et \textit{reproductibilité}, entre \textit{stabilité} et \textit{accessibilité}, ou encore entre \textit{qualité des résultats} et \textit{frugalité computationnelle}.' -> '...frugalité calculatoire.'; 'chaque fois que cela était possible, au prix d’une perte de fidélité' -> 'chaque fois que possible, quitte à perdre en fidélité'
Certaines métriques récentes, telles que \textit{G-Eval} reposant sur GPT-4, dépendent de modèles ou d’API fermées, ce qui soulève plusieurs enjeux critiques pour la reproductibilité :
\begin{itemize}
    \item un \textbf{biais d’accessibilité}, lié aux ressources financières ou matérielles nécessaires pour interroger ces modèles ;
    \item des \textbf{limitations techniques}, telles que les quotas d’utilisation, les coûts d’exécution ou la latence ;
    \item une \textbf{fragilité de la reproductibilité} à moyen terme, certains modèles pouvant être modifiés ou rendus inaccessibles (comme GPT-3), ce qui remet en question la pérennité des expériences.
\end{itemize}
Ces contraintes illustrent des points de tension : entre \textit{performance} et \textit{reproductibilité}, entre \textit{stabilité} et \textit{accessibilité}, ou encore entre \textit{qualité des résultats} et \textit{frugalité calculatoire}. Dans notre étude, nous avons privilégié l’utilisation d’alternatives open-source, quitte à perdre en fidélité par rapport aux conditions d’évaluation originales.

\paragraph{Données d’évaluation limitées à un seul benchmark.}
% Correction: '\textit{SummEval}, qui bien que riche.' -> '\textit{SummEval}, bien que riche.'; 'généraliser nos conclusions. Notre framework a été codé pour faciliter l'utilisation de nouveaux jeux de données.' -> 'généraliser nos conclusions. Notre framework a cependant été conçu pour faciliter l’ajout de nouveaux jeux de données.'
Notre étude repose exclusivement sur le jeu de données \textit{SummEval}, bien que riche. Des expérimentations sur d’autres corpus seraient nécessaires pour généraliser nos conclusions. Notre framework a cependant été conçu pour faciliter l’ajout de nouveaux jeux de données.

\paragraph{Temps de calcul prohibitif pour certaines métriques.}
% Correction: '\textit{SEval-Ex}, \textit{QuestEval} ou \textit{G-Eval},' -> Ok; 'très élevé (jusqu’à plusieurs heures),' -> 'très élevé (jusqu’à plusieurs heures par exécution),'
Certaines métriques, comme \textit{SEval-Ex}, \textit{QuestEval} ou \textit{G-Eval}, présentent un temps de traitement très élevé (jusqu’à plusieurs heures par exécution), les rendant peu adaptées aux environnements à contrainte de ressources ou aux évaluations à grande échelle.

\section{Conclusion}\label{sec:conclusion}
% Correction: 'soulignent l’urgence d’une approche plus rigoureuse, explicite et systématique' -> 'soulignent l’urgence d’adopter une approche plus rigoureuse, explicite et systématique'; 'frugalité computationnelle' -> 'frugalité calculatoire'; '(comme G-Eval ou SEval-Ex)' -> Ok; 's’avère parfois incompatible avec la reproductibilité :' -> 's’oppose parfois à la reproductibilité :'; 'pertinence évaluative :' -> 'pertinence évaluative :'; 'ROUGE,' -> Ok; 'modèles modernes.' -> 'modèles modernes.'; 'plateforme unifiée pour la comparaison transparente et équitable' -> 'plateforme unifiée facilitant la comparaison transparente et équitable'; '(modèles, prompts, températures, seeds, etc.),' -> '(modèles, prompts, températures, graines aléatoires (seeds), etc.),'; 'rapports de performance.' -> 'rapports de performance.'; 'évaluation automatique se déplace vers des approches fondées sur les LLMs,' -> 'évaluation automatique s’oriente vers des approches fondées sur les LLM,'; 'compréhension des outils existants, mais aussi la conception de métriques nouvelles,' -> 'compréhension des outils existants mais aussi la conception de nouvelles métriques,'; 'efficience computationnelle.' -> 'efficacité calculatoire.'
Notre étude met en lumière les défis majeurs que pose la reproductibilité dans l’évaluation automatique des résumés. Les écarts parfois significatifs entre les performances rapportées dans la littérature et celles mesurées au sein de notre cadre expérimental soulignent l’urgence d’adopter une approche plus rigoureuse, explicite et systématique dans ce domaine.

Trois tensions structurantes émergent de nos analyses. Premièrement, la performance s’oppose souvent à la frugalité calculatoire : les métriques les plus corrélées aux jugements humains (comme G-Eval ou SEval-Ex) sont également les plus coûteuses en ressources. Deuxièmement, la performance s’oppose parfois à la reproductibilité : certaines métriques, bien que prometteuses, montrent une variabilité préoccupante selon le modèle ou l’environnement d’exécution. Troisièmement, la simplicité d’usage entre en conflit avec la pertinence évaluative : les métriques classiques telles que ROUGE, faciles à implémenter, peinent à capturer la richesse des résumés produits par des modèles modernes.

Notre framework open-source constitue une réponse partielle à ces défis en offrant une plateforme unifiée facilitant la comparaison transparente et équitable des métriques. Pour aller plus loin, nous recommandons à la communauté d’adopter des protocoles standardisés, de publier systématiquement les paramètres expérimentaux complets (modèles, prompts, températures, graines aléatoires (seeds), etc.), et d’intégrer des mesures statistiques de variabilité dans les rapports de performance.

Ces efforts de standardisation sont d’autant plus essentiels que l’évaluation automatique s’oriente vers des approches fondées sur les LLM, dont le comportement intrinsèquement stochastique accentue les risques de non-reproductibilité. À long terme, de telles pratiques favoriseront non seulement une meilleure compréhension des outils existants mais aussi la conception de nouvelles métriques, capables de concilier exigence méthodologique, robustesse empirique et efficacité calculatoire.
% Le paragraphe commenté suivant semble redondant avec la conclusion finale et peut être supprimé.
% % Ces travaux mettent en lumière une tension entre performance, reproductibilité, stabilité et frugalité. L'utilisation de modèles puissants peut améliorer la corrélation avec les jugements humains, mais au prix d'une reproductibilité fragile, de sorties instables, et de coûts computationnels considérables. Il est donc impératif de mieux documenter, quantifier et anticiper ces phénomènes dans la conception des métriques d’évaluation fondées sur les LLMs.

%%================================================================
%% Note : si l'on préfère éviter de factoriser les crossrefs :
%% bibtex -min-crossrefs 99 taln-exemple
%%================================================================
\bibliographystyle{coria-taln2025}
\bibliography{biblio} % Assurez-vous que le fichier biblio.bib est correct et complet.
% \nocite{TALN2015,LaigneletRioult09,LanglaisPatry07,SeretanWehrli07} % À supprimer pour la version finale sauf si ces références doivent absolument apparaître.

%%================================================================
\end{document}